\documentclass[a4paper,conference]{IEEEtran}
\ifCLASSINFOpdf
\else
\fi
\usepackage[table]{xcolor}

\usepackage{amsmath}
\usepackage{amssymb}
\usepackage{bm}
\usepackage{svg}
\usepackage{graphicx}
\usepackage{blindtext}
\usepackage{array}
\usepackage{etoolbox}
\usepackage{wrapfig}
\usepackage{color,soul}
\usepackage{multirow}
\usepackage{threeparttable}
\usepackage{mathabx}
\usepackage{tabularx}
\usepackage{lipsum}

\graphicspath{{figures/}}

\begin{document}
\title{Skeletal Human Action Recognition using Hybrid Attention based Graph Convolutional Network}

\author{\IEEEauthorblockN{Hao Xing\IEEEauthorrefmark{1},
Darius Burschka\IEEEauthorrefmark{2}}
\IEEEauthorblockA{Technical University of Munich\\
 Machine Vision and Perception Group\\
 Munich Institute of Robotics and Machine Intelligence, Department of Computer Science\\
 Parkring 13, 85748, Munich, Germany}
 Email: \IEEEauthorrefmark{1} hao.xing@tum.de,
\IEEEauthorrefmark{2} burschka@cs.tum.edu
}


%Telephone: (800) 555--1212, Fax: (888) 555--1212}


\maketitle


\begin{abstract}
    In skeleton-based action recognition, Graph Convolutional Networks model human skeletal joints as vertices and connect them through an adjacency matrix, which can be seen as a local attention mask. However, in most existing Graph Convolutional Networks, the local attention mask is defined based on natural connections of human skeleton joints and ignores the dynamic relations for example between head, hands and feet joints. In addition, the attention mechanism has been proven effective in Natural Language Processing and image description, which is rarely investigated in existing methods. In this work, we proposed a new adaptive spatial attention layer that extends local attention map to global based on relative distance and relative angle information.  Moreover, we design a new initial graph adjacency matrix that connects head, hands and feet, which shows visible improvement in terms of action recognition accuracy. The proposed model is evaluated on two large-scale and challenging datasets in the field of human activities in daily life: NTU-RGB+D and Kinetics skeleton. The results demonstrate that our model has strong performance on both dataset.
\end{abstract}

% \fi
\IEEEpeerreviewmaketitle

\section{Introduction}
\label{sec:intro}
Action recognition using skeletal information has been widely investigated and attracted a lot attention, since most of human body can be viewed as an articulated system with rigid bones connected by joints, which are not sensitive to the background and the appearance of human \cite{shi2019two, yan2018spatial}. 

Most existing methods are using Convolutional Neural Networks (CNNs) and Recurrent Neural Networks (RNNs) to model respectively human skeleton spatial structure and temporal dynamics. However, both cannot fully represent the spatial and temporal features of human skeleton at the same time \cite{yan2018spatial}. In these networks, the skeleton input is usually processed as a pseudo-image or sequence of joint coordinate vectors, ignoring the spatial connections between the skeleton joints \cite{shi2019two}. RNNs are additionally limited by the short-memory in analyzing global temporal features. Moreover, it is hard to generalize the graph structure of skeleton data to any random form of skeleton using previous methods. 

Recently, with development of Graph Convolutional Networks (GCNs), a compatible solution is proposed. In GCNs, the spatial features can be represented by a spatial graph that is the combination of joints (vertices) and their natural connection (edges). With similar definition, the temporal features can be demonstrated by a temporal graph that connects each vertex and its neighbors in consecutive frames with temporal edges \cite{yan2018spatial}. Normally, these spatial and temporal edges are defined by the natural connections, such as connections of the elbow to the wrist and the shoulder, which are the same for different actions. However, it is not suitable for the body-parts related activity for example drinking and eating, which has strong relations between different body parts, like hands and head. In order to extract the dynamic relations of different actions, an adaptive mechanism is in demand. 

In the field of Natural Language Processing (NLP), the attention mechanism is successfully applied to find the potential relations between words at different positions \cite{vaswani2017attention}. For a similar purpose, we propose a hybrid spatial attention mechanism in GCN to generate new edges between strongly related vertices during training process, which automatically adapts to different graph description of actions and different input streams.

% With an adaptive mechanism, the spatial graph (adjacency matrix) can be updated during training \cite{shi2019two} and generate the new edge between strongly related vertices. However, the adaptive mechanism did not fully use the attention mechanism, which achieved great success in Nature Language Processing (NLP) and image description.

Overall, the main technical contributions of our work lie in three fields:
\begin{itemize}
    \item[1.] We design a new spatial graph with connection between head, hands and feet.
    \item[2.] We propose a novel adaptive mechanism that uses a spatial hybrid attention (HA) layer with mixture of relative distance and relative angle information, in which the relative distance attention contributes more to the bone stream related action recognition and the relative angle attention is more beneficial to the joint stream related action classification.
    \item[3.] We evaluate our model on two large-scale and challenging datasets in the field of human action recognition: NTU-RGB+D and Kinetics Skeleton, and our model achieves a strong performance on both datasets. 
\end{itemize}

The rest of the paper is organized as follows: in section \ref{sec:2}, we briefly review existing approaches of human action recognition using skeletal information, graph convolutional network and attention based network. Section \ref{sec:3} introduces our proposed Attention Based Graph Convolutional Network. Section \ref{sec:4} reports experimental results and discussions. Section \ref{sec:5} concludes the paper.
	
% 1) We design a new spatial graph with connection between head, hands and feet. 2) We propose a novel adaptive mechanism that uses a new spatial attention (SA) along with a channel attention (CA) layer. 3) We evaluate our model on two large-scale and challenging datasets in the field of human action recognition: NTU-RGBD and Kinetics Skeleton, both achieve a strong performance on action recognition, in which the SA layer contributes more to the body parts related action recognition and the CA layer is more benefit to the pose related action classification.

\section{Related Work}
\label{sec:2}
We review the previous works from three primary related streams of the research area: skeleton based human action recognition, graph convolution networks, and attention based convolutional networks.

\subsection{Skeleton based human action recognition}
% \paragraph{Skeleton based human action recognition}
Human action recognition using skeletal information can be categorized into two clusters: handcrafted feature based methods and learning based methods. The first method manually designs several features to model human body. Vemulapalli \emph{et al.}~\cite{vemulapalli2014human} represented the action sequence as a curve in Lie group \textit{SE}(3) $\times \dots \times$ \textit{SE}(3), which can be mapped into its Lie algebra and form a feature vector.
% Hussein \emph{et al.}~\cite{hussein2013human} constructed a matrix descriptor according to covariance of each joints. 
Fernando \emph{et al.}~\cite{fernando2015modeling} adopted a ranking machine to extract the appearance feature changing of frames evolves with time. In our previous work \cite{xing2021robust}, we modeled the fall down event with spatial action unit and temporal height change of skeleton, which has a good performance on single event detection. However, these methods are barely satisfied for large-scale multi-class action recognition, because of the complexity of action space. 

Recently, with the success of data driven methods, Deep Learning methods have been widely applied in the field of human action recognition. These approaches are mostly using Convolution Neural Networks (CNNs) and Recurrent Neural Networks (RNNs). CNN-based methods convert each skeleton frame to a pseudo image using designed transformation strategy \cite{li2021memory}. Baradel \emph{et al.}~\cite{baradel2017human} combined human skeletal information with RGB images, which can offer richer contextual cues for action recognition. The authors introduced also a spatial hands attention mechanism, which crops the image around hands. RNN-based methods emphasize the temporal dynamics of skeleton joints \cite{wang2017modeling}. Zhang \emph{et al.}~\cite{zhang2017view} proposed an adaptive RNN model, which can adjust to the most suitable observation viewpoints for cross-view action recognition. Si \emph{et al.}~\cite{si2019attention} reformed the input skeleton information into the graph-structured data through a graph convolutional layer within the Long Short-Term Memory (LSTM) network. 

However, both CNN-based and RNN-based methods cannot fully model human action spatial features and temporal features \cite{yan2018spatial}. Both methods ignore the spatial connections between joints, and RNN-based methods suffer from short-memory in analyzing global temporal features.

\subsection{Graph convolution networks}
% \paragraph{Graph convolution networks}

Recently, Graph Convolution Networks (GCN) designed for structured data representation raise the attention. The Graph Convolution Networks (GCNs) can also be categorized into two clusters: spatial and spectral. The spatial GCNs operate the graph convolutional kernels directly on spatial graph nodes and their neighborhoods \cite{shi2019skeleton}. Yan \emph{et al.}~\cite{yan2018spatial} proposed a Spatial-Temporal Graph Convolutional Network (ST-GCN), which extract spatial feature from the skeleton joints and their naturally connected neighbors and temporal feature from the same joints in consecutive frames. Shi \emph{et al.}~\cite{shi2019two} introduced a two stream Adaptive Graph Convolutional Network (2s-AGCN) based on ST-GCN, which not only extracts features from skeleton joints but also considers the direction of each joint pair (bone information). 

The spectral GCNs consider the graph convolution in form of spectral analysis \cite{li2015gated}. 
Henaff \emph{et al.}~\cite{henaff2015deep} developed a spectral network incorporating with graph neural network for the general classification task. Kipf and Welling \cite{kipf2016semi} extends the spectral convolutional network further in the field of semi-supervised learning on graph structured data.

This work follows the spatial GCNs that apply the graph convolutional kernels on spatial domain. 

\subsection{Attention based convolutional networks}
% \paragraph{Attention based convolutional networks}

Attention based neural networks have been successfully applied in NLP and image description. In the field of NLP, the multi head self-attention layer generates the representation of a sequence by aligning words in the sequence with other words \cite{velivckovic2017graph}. Vaswani \emph{et al.}~\cite{vaswani2017attention} employed a local attention mechanism on each node and its neighbor pairs in parallel, so that the spatial feature from each neighbor node is weighted by the relative relationship. Devlin \emph{et al.}~\cite{devlin2018bert} extended a self-attention layer bidirectionally, which can model many downstream tasks in text processing.

In the field of image description, the attention mechanism is adopted to generate a learnable weight mask in spatial domain, which demonstrates the importance of a region \cite{xu2015show}. Liu \emph{et al.}~\cite{liu2017attention} adopted an attention correctness mechanism to generate the attention mask for a corresponding image area. Anderson \emph{et al.}~\cite{anderson2018bottom} combined a top-down attention based CNN with a bottom-up Fast R-CNN to determine feature weightings for each detected region. 

As part of natural language, human actions also have strong attention relations between different body parts, such as relative distances and relative angles. Inspired by aforementioned great previous works, we attempt to improve the performance of the graph convolutional networks on Human Action Recognition by designing a novel attention mechanism.

% Considering human action as part of Natural Language,  we decide to develop an attention mechanism based Graph Convolutional Network for large-scale human action recognition.

\section{Hybrid Attention based Graph Convolutional Network}
\label{sec:3}
% \subsection{Four stream of information}
Typically, the raw skeleton data are provided as a sequence of vectors. Each vector contains a set of human joint coordinates in 2D or 3D. Then a bone is defined by the difference of its two ends joints. In a graph representation, the joint and bone (spatial) information can be viewed as vertices and their natural connections are edges as shown in Fig \ref{fig:graph} (a). Besides joint and bone information, we also generate their velocity (temporal) information, and feed them together with spatial information into the attention based graph convolutional network (AB-GCN)

% we add additional edges between wrists, feet and head to the natural connections, since.

\subsection{Graph construction}

\begin{figure}
% \begin{wrapfigure}{l}{1.0\linewidth}
    % \vspace*{-3\baselineskip}
    \begin{tabular}{@{}cc@{}}
    \includegraphics[width=0.24\textwidth]{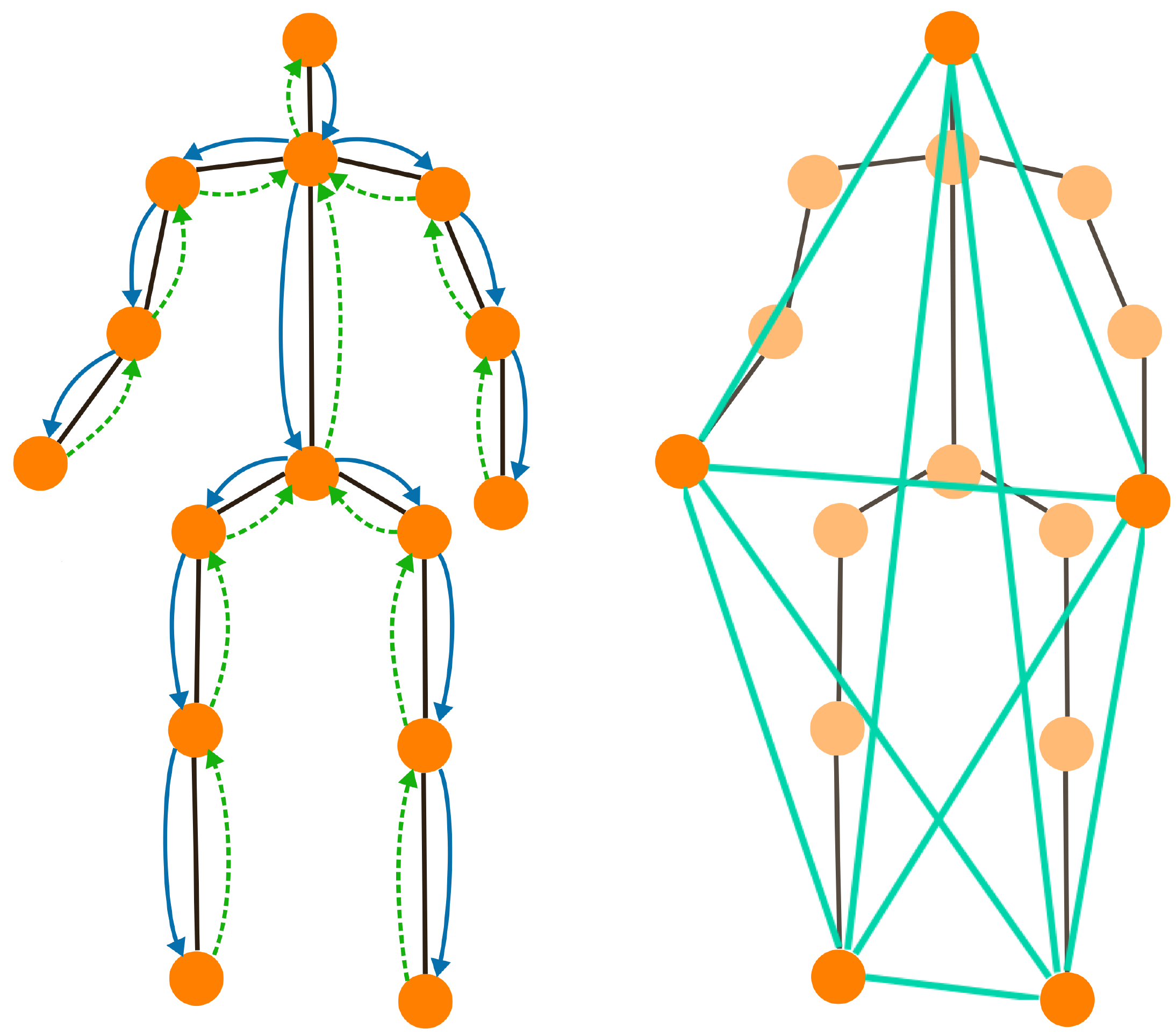} &% Dummy image replacement
    \includegraphics[width=0.203\textwidth]{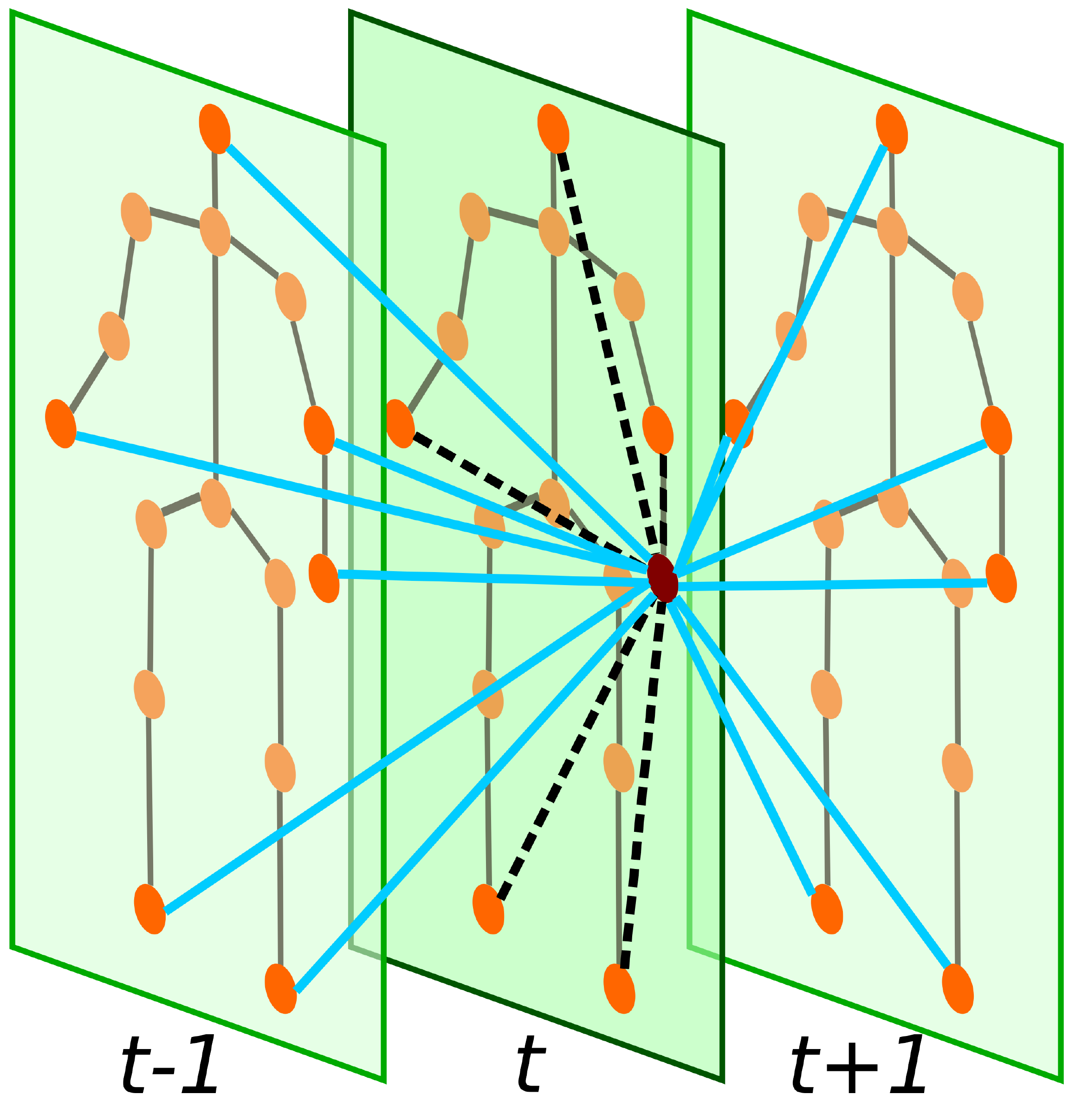} \\% Dummy image replacement
    (a) & (b)
    \end{tabular}
  \caption{\label{fig:graph}Illustration of skeleton graph: (a) Left: Spatial graph with notes and edges (solid arrow: outward edges; dash arrow: inward edges); Right: Additional connections between head, hands and feet; (b) Temporal (solid blue) and spatial (dash black) edges of the left wrist node.}
  \vspace*{-1.\baselineskip}
% \end{wrapfigure} 
\end{figure}

\begin{figure*}[t]
\centering
\begin{tabular}{c}
    \includegraphics[width=0.98\textwidth]{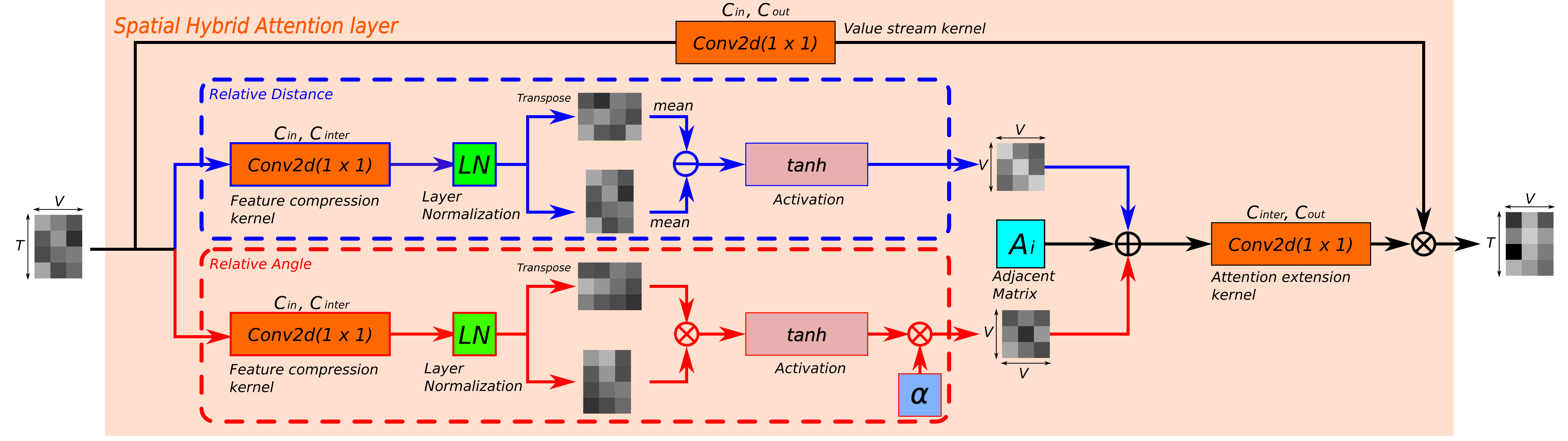} 
\end{tabular}
\caption{Illustration of the spatial hybrid attention convolution layer, where blue stream (top) is ``\textit{Relative Distance}" attention branch and red stream (bottom) represents `\textit{Relative Angle}" attention branch. $C_{in}$, $C_{inter}$ and $C_{out}$ stand for input, inter and output channel, respectively. $V$ and $T$ represent the spatial and temporal size of feature (attention) map. ``\textit{mean}" is an average process in temporal dimension.}
\label{fig:attention}
\vspace*{-0.5\baselineskip}
\end{figure*}

The traditional skeleton graph follows the work of ST-GCN~\cite{yan2018spatial}, which constructs a graph using the natural structure of the human body as shown in Fig \ref{fig:graph} (a) left. However, it ignores the strong relations between the parts that usually have large movements, such as hands, head and feet.

Hence, we add additional connections between those parts as shown in Fig \ref{fig:graph} (a) right. Each connection has the same incoming, outgoing and self-connecting edges as the traditional graph. All edges form a binary adjacency matrix $\mathbf{A}_{init}$, in which $a_{ij} = 1$ means vertices $v_i$ and $v_j$ are connected. Then a spatial graph can be mathematically expressed as: 
\begin{equation}
    \mathbf{G} = \mathbf{A}\cdot\mathbf{F}_{in}
\end{equation}
where $\mathbf{F}_{in}$ is the input skeleton feature map, $\mathbf{G}$ is the graph feature map and $\mathbf{A}$ is column-wise normalization of $\mathbf{A}_{init}$.
The temporal graph is constructed by connecting vertices and their neighbor pairs in consecutive frames in the same way, as shown in Fig. \ref{fig:graph} (b).

% \begin{wrapfig}{r}{0.5\textwidth}
% \begin{figure*}
% \begin{centering}
% \begin{tabular}{cc}
% \fbox{\includegraphics[width=0.24\textwidth]{graph_construction_both_version2}}&
% \fbox{\includegraphics[width=0.203\textwidth]{graph_construction_temporal_version2}}\\
% (a)&(b)
% \end{tabular}
% \caption{Illustration of skeleton graph: (a) Left: Spatial graph with notes and edges (solid arrow: outward edge; dash arrow: inward edge); Right: Additional connections between head, hands and feet; (b) Temporal (solid blue) and spatial (dash black) edges of one node connect the node with its neighborhoods in consecutive frames}
% \label{fig:graph}
% \end{centering}
% \end{figure*}
% \end{wrapfig}

\subsection{Graph convolutional layer}

Given the defined skeleton graph with size $C \times T \times V$, where $C$ denotes number of channels, $T$ is number of frames and $V$ is vertices volume, we apply graph convolutional layer on spatial and temporal dimension to extract distinguishable features. The graph convolutional layer has two types: spatial layer and temporal layer. 

In spatial dimension, the layer operates the vertices with a $1\times 1$ convolutional kernel as follows:

\begin{equation}
    \mathbf{F}_{out} = \sigma(\mathbf{W}_s\mathbf{G}_{in}+ \mathbf{B}_s) = \sigma(\mathbf{W}_s \mathbf{A} \mathbf{F}_{in}+ \mathbf{B}_s)
\end{equation}
where $\mathbf{F}_{out}$ denotes output feature map, $\mathbf{G}_{in}$ means input graph feature,
% fixed by the graph construction according to
$\mathbf{W}_s\in \mathbb{R}^{C_{out}\times C_{in}\times 1\times 1}$ and $\mathbf{B}_s\in\mathbb{R}^{C_{out}\times C_{in}\times 1\times 1}$ are parameters of the spatial convolutional kernel, $C_{out}$ and $C_{in}$ are the output and input channel number, $1\times 1$ indicates the kernel size, 
and $\sigma$ is the \textit{ReLU} activation function

In temporal dimension, the temporal layer is applied with four dilated $k \times 1$ convolutional kernels. Note that the kernel size in spatial dimension ($V$) is fixed at $1$, because the spatial relationship is previously defined by the graph construction. 

It can be seen from this point that the initial adjacency matrix has a significant influence on final prediction result. In ST-GCN~\cite{yan2018spatial}, Yan \emph{et al.} used a fixed adjacency matrix, which means that it can only update the importance of existing edge and cannot create a new edge during training process. Different from ST-GCN, Shi \emph{et al.}~\cite{shi2019two} developed an adaptive GCN (2s-AGCN), which set the adjacency matrix as a parameter and update it by the spatial attention layer during training. Inspired by their job, we found that the adjacency matrix is a given local attention map that is based on the prior knowledge. Hence, an attention layer can be added into a GCN to extend the local attention to global. In doing so, we design a hybrid attention based GCN (HA-GCN) and try to improve the performance of GCN on skeletal action recognition by exploring combination of different spatial attentions, i.e. relative distance and relative angle.

% \begin{figure}[t]
% \begin{tabular}{c}
% % \fbox{\includegraphics[width=0.95\textwidth]{spatial_attention.eps}}\\
% \includegraphics[width=0.95\textwidth]{spatial_attention.eps}\\
% (a) \\
% \includegraphics[width=0.94\textwidth]{channel_attention.eps} \\
% % \fbox{\includegraphics[width=0.95\textwidth]{channel_attention.eps}}\\
% (b)
% \end{tabular}
% \caption{Illustration of attention unit: (a) Spatial convolution unit with spatial attention unit; (b) Channel attention unit ($C_{in} = C_{out}$);}
% \label{fig:attention}
% \end{figure}

\subsection{Hybrid attention graph convolution layer}
\label{method:attention_layer}

\begin{figure*}
\begin{tabular}{cc}
\fbox{\includegraphics[width=0.45\textwidth]{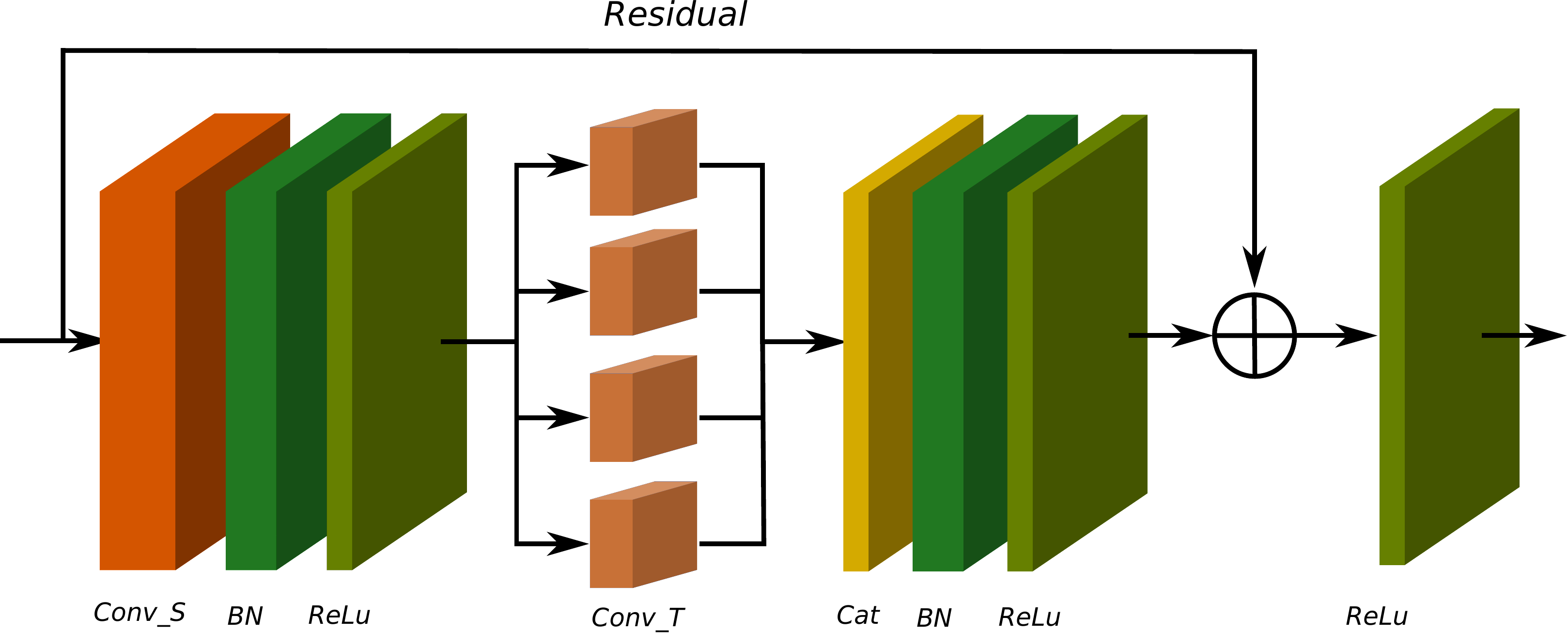}}
 &
\fbox{\includegraphics[width=0.49\textwidth]{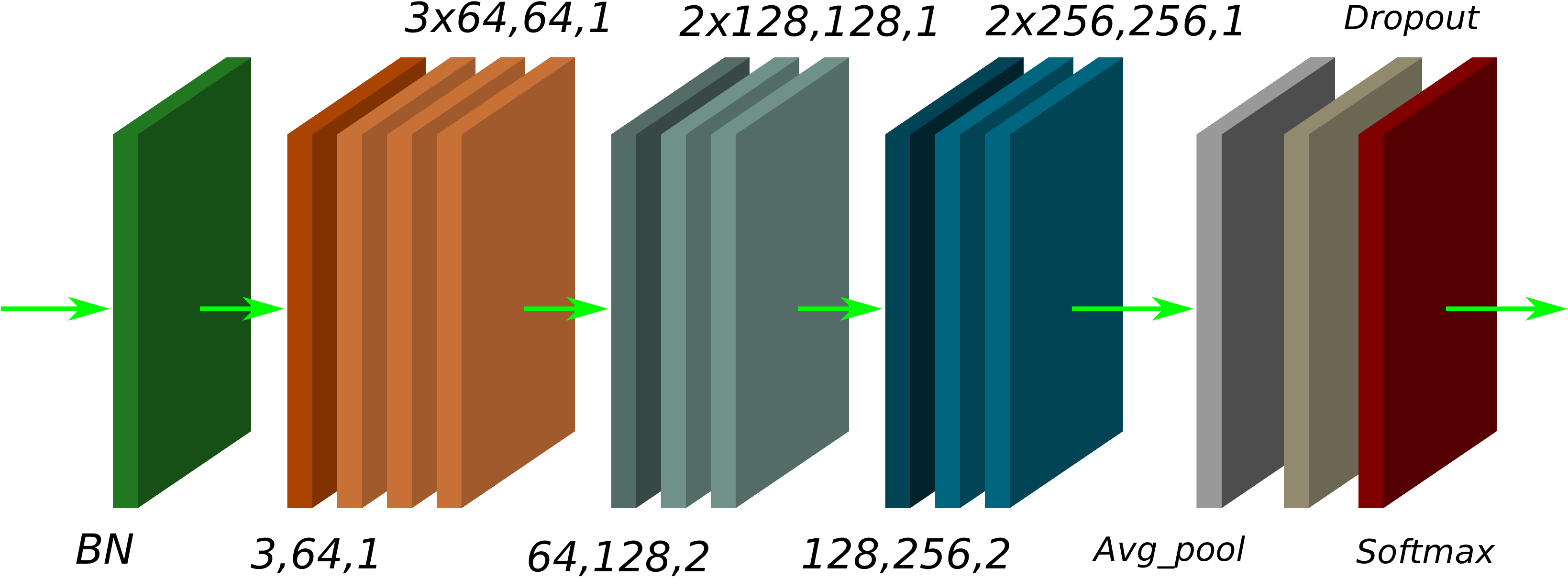}}\\
(a) & (b)\\
\end{tabular}
\caption{Illustration of hybrid attention based graph convolutional network: (a) Hybrid attention based graph convolutional block unit consisting of spatial convolutional layer (\textit{Conv\_S}), \textit{Batch Normalization} (\textit{BN}), temporal convolutional layer (\textit{Conv\_T}), concatenate function (\textit{Cat}) and \textit{ReLU} activation function; (b) Hybrid attention based graph convolutional network that consists of 10 HA-GCN blocks, where the input channel, output channel and stride parameters are listed besides the block, such as $3,64,1$ mean $3$ input channel, $64$ output channel, and $1$ stride, respectively, $2\times$ and $3\times$ represent $2$ and $3$ same blocks, and \textit{Avg\_pool} is the average pooling function.}
\label{fig:block}
\vspace*{-1.5\baselineskip}
\end{figure*}
\begin{figure}
\resizebox{\linewidth}{!}{
\begin{tabular}{cc}
\includegraphics[width=0.42\textwidth]{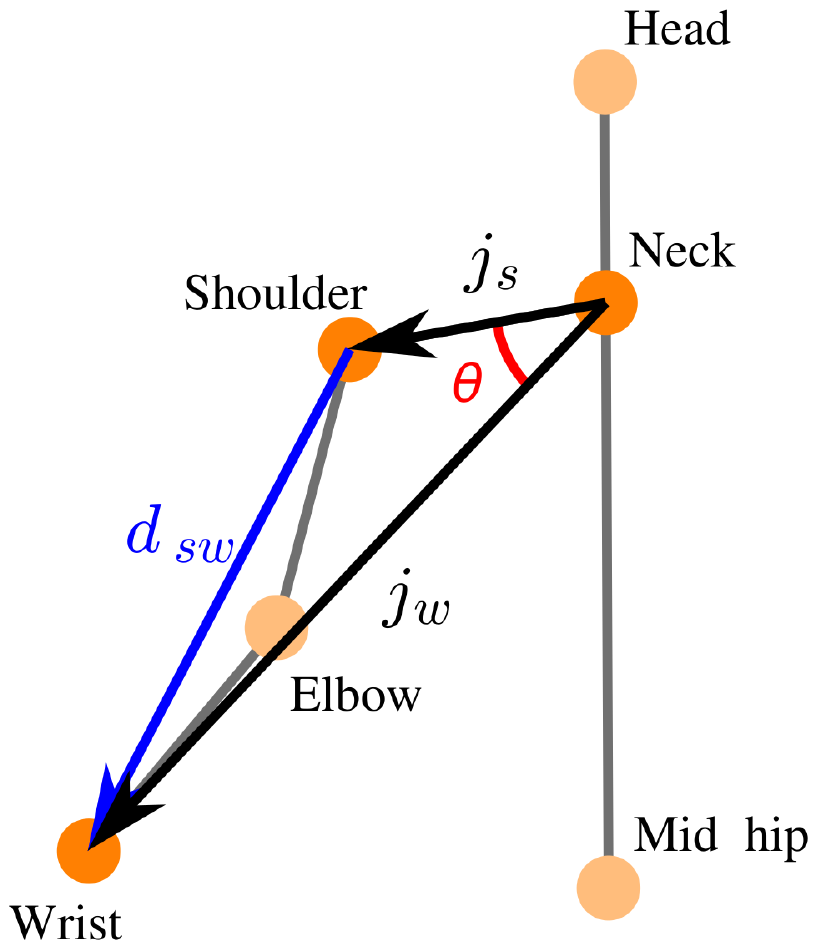}
 &
\includegraphics[width=0.45\textwidth]{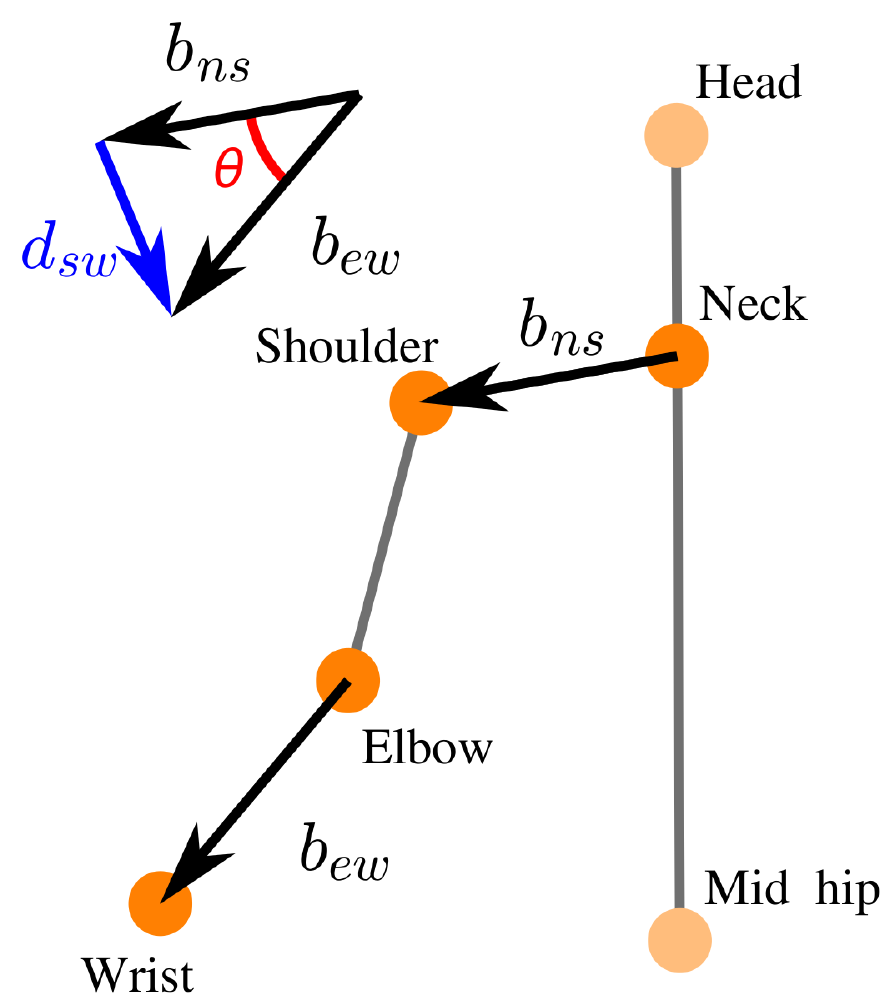}\\
(a) Joint stream & (b) Bone stream\\
\end{tabular}
}
\caption{Examples of Relative Distance and Relative Angle in the spatial domain over the (a) Joint and (b) Bone streams. In the joint stream (a), the origin point is \emph{Neck}, ${\bm{j}}_s$ and ${\bm{j}}_w$ are the joint vectors of \emph{Shoulder} and \emph{Wrist}. In the bone stream, ${\bm{b}}_{ns}$ and ${\bm{b}}_{ew}$ are bone connections of \emph{Neck}-\emph{Shoulder} and \emph{Elbow}-\emph{Wrist}. The $\theta$ is the relative angle, the $\bm{d}_{sw}$ is the relative distance}
\label{fig:RARD}
\vspace*{-1.5\baselineskip}
\end{figure}

% Recently, attention mechanism achieve great success in NLP and image description generation. 
The attention layer relates different features of a same input and generates a mask map that contains the importance of each element in feature map. The importance (score) can be expressed as follows:
\begin{equation}
    \label{eq:dot_production_attention}
    m_{ij} = \text{score}({\bm{f}}_i, \bm{f}_j) = \frac{\bm{f}_i^T \bm{f}_j}{\sqrt{n}}
\end{equation}
where $m$ is an element of mask map $\mathbf{M}$, $\bm{f}_i$ and $\bm{f}_j$ are elements in the feature map, and $n$ is a normalizing parameter, it can be the length of a vector, when $\bm{f}_i$ and $\bm{f}_j$ are column feature vectors. Given such a mask map, typically, a \emph{softmax} function is applied to normalize the scores into range $\left[0,1\right]$. In this work, we find that feeding mask maps into a 2-dimensional convolution layer contributes to the relations learning process.  

In spatial dimension, we follow the three types spatial graph structure from 2s-AGCN~\cite{shi2019two}, which are respectively generated by identity, inwards and outwards adjacency matrix. On each graph, we apply a new designed hybrid attention layer to extract spatial attention information. The hybrid attention consists of two branches: Relative Distance (RD) attention and Relative Angle (RA) attention, which have significant benefit for bone stream and joint stream, respectively. The proof can be found in Sec~\ref{sec:4} Experiments. 

For both attention branches, the input feature map will be first compressed in channel dimension by a 2D convolutional kernel. In doing so, it reduces the computation of attention, increases the feature difference between channels, and further can generate an unique attention map for each channel. In this work, the compression ratio is $C_{inter}/C_{in}=8$. In order to generate stable attention for different distribution of input action cases, a Batch Normalization (BN) function is usually used before calculating the attention score. However, this damages the performance of the model with small-batch size as it does not contain a representative distribution of examples \cite{ioffe2017batch}. Therefore, we adopt the Layer Normalization (LN) function to allow each input case standardize only on its own batch. As shown in Fig.~\ref{fig:attention}, the compressed feature map is normalized by the \textit{Layer Normalization} function for each batch, and is fed into the corresponding attention function together with its transpose feature map. This process can be formulated as follows:

\begin{equation}
    \mathbf{F} = Ln(\mathbf{W}_c\mathbf{X} + \mathbf{B}_c)
\end{equation}
where $\mathbf{W}_c\in \mathbb{R}^{1\times 1}$ and $\mathbf{B}_c\in\mathbb{R}^{1\times 1}$ are parameters of the feature compression kernel, \textit{Ln} is the \textit{Layer Normalization} function, and $\mathbf{X}$ are the input feature map, and $\mathbf{F}$ are the compressed feature map.

Since the attention score is calculated between nodes, we take two node feature vectors $\bm{f}_i$ and $\bm{f}_j$ from the compressed feature map $\mathbf{F}$ as examples to further explain our attention mechanism, they are $1\times T$ in size.

\noindent
\textit{Relative Distance attention}: The RD attention information is generated by the relative distance between nodes as follows:

\begin{equation}
    a_{RD,ij} = \text{tanh}((\bar{f}_i - \bar{f}_j)),\ \text{with}\ i,j\in[1, V] 
\end{equation}
where $a_{RD}$ is an element of RD attention mask $\mathbf{A}_{RD}$, \textit{tanh} is the \emph{Hyperbolic Tangent} activation function, $\bar{f}$ is the average value of feature vector $\bm{f}$ over temporal dimension, $i, j$ are the indices of nodes. The final RD attention mask $\mathbf{A}_{RD}$ is with size of $V\times V\times C_{inter}$, where $C_{inter}$ is the number of channel.

\noindent
\textit{Relative Angle attention}: The RA attention information is obtained by the dot product between node feature vectors in channel-wise, as demonstrated:
\begin{equation}
\begin{split}
    a_{RA,ij}& = \text{tanh}(\bm{f}^T_i\cdot \bm{f}_j)\\
    &=\text{tanh}(|\bm{f}_i||\bm{f}_j|cos(\theta))
    ,\ \text{with}\ i,j\in[1, V] 
\end{split}
\label{eq:RA}
\end{equation}
where the $\theta$ is the angle between two vectors. Note that we simplify the dot product attention in Eq.~\ref{eq:dot_production_attention} by removing the scale $1/\sqrt{n}$, since $n$ is the number of nodes and it is constant. The Eq.~\ref{eq:RA} will generate an RA attention mask of the same size as the RD mask. As can be seen from the examples in Fig.~\ref{fig:RARD}, relative distance and relative angle focus on different features on joint and bone streams. In some actions, such as drinking, eating, etc., the vector pairs with small relative distance (\emph{Head}-\emph{Wrist}) should have a great influence on the action prediction. At this time, the relative angle mechanism will draw the attention to these pairs, due to $cos\theta \approx 1$. In some other actions, e.g., stretching, celebrating and so on, the relative distance attention should dominate the attention value.

Since the two attention mechanism (RA and RD) have different effects on different actions, we adopt a learnable parameter $\alpha$ to combine them. The sum hybrid attention score $A_{h}$ are formed by the following equation:
\begin{equation}
    \mathbf{A}_{h} = \mathbf{A}_{RD} + \alpha\cdot \mathbf{A}_{RA}
\end{equation}

% As shown in the Fig \ref{fig:attention}, the input feature map goes into two parallel branches. In each branch, it is first fed into one 2D convolution kernel and normalized for each batch by a Layer Normalization function. The output feature map is transposed and subtracted or multiplied by itself to generate RD and RA attention information, respectively. 
As aforementioned, the predefined adjacency matrix is a local attention map, and the spatial attention mechanism is adaptive to different input action classes, which can enrich the local attention into a global map. Hence, the final attention map is generated by the combination of the hybrid attention mask with the adjacent matrix as follows:

\begin{equation}
    \mathbf{A}_{final} = \mathbf{A}_h + \mathbf{A}_i = \mathbf{A}_{RD} + \alpha\cdot \mathbf{A}_{RA}+\mathbf{A}_i
\end{equation}
Note that the initial graph mask $\mathbf{A}_i$ is added in channel-wise, since it is of size $V\times V \times 1$, while the size of the attention masks are $V\times V \times C_{out}$.
% \hl{explain why LN no BN and why single conv} 

% Their matrix production is then fed into a 1D convolution layer with a \emph{sigmoid} activation function to extract the attention mask. As aforementioned, the predefined adjacency matrix is a local attention map and has a guiding effect on the initial attention. Hence, the final attention map is generated by the combination of the attention mask with the adjacent matrix as follows:
% \begin{equation}
%     Attention = M + A_i = Conv1d(F_1^T\cdot F_2)+A_i
% \end{equation}
% where $M$ is the attention mask that is extracted by the 1D convolution layer on the production of feature maps $F_1$ and $F_2$, and $A_i$ is one of the three adjacent matrices. In order to give more flexibility to the spatial graph, we set adjacency matrices as learnable parameters with given initial values. 

The final attention map is extended to $C_{out}$ output channels through an additional 2D convolution layer and merged with the value stream by matrix multiplication. The process can be mathematically expressed as follows:
\begin{equation}
    \mathbf{Y} = \sigma((\mathbf{W}_{A}\mathbf{A}_{final}+ \mathbf{B}_{A})*(\mathbf{W}_{v}\mathbf{X}+ \mathbf{B}_{v}))
\end{equation}
where $\mathbf{W}_A\in \mathbb{R}^{C_{out} \times C_{inter} \times 1\times 1}$ and $\mathbf{B}_A\in\mathbb{R}^{C_{out} \times C_{inter} \times 1\times 1}$ are parameters of the attention extension kernel, $\mathbf{W}_v\in \mathbb{R}^{C_{out} \times C_{in} \times 1\times 1}$ and $\mathbf{B}_v\in\mathbb{R}^{C_{out} \times C_{in} \times 1\times 1}$ are parameters of the value stream kernel, $\mathbf{X}$ is the input feature map, and $\mathbf{Y}$ is the output feature map.

% \begin{wrapfigure}{r}{0.7\textwidth}
% \centering
% % \begin{tabular}{c}
% % \fbox{\includegraphics[width=0.95\textwidth]{spatial_attention.eps}}\\
% \includegraphics[width=0.7\textwidth]{channel_attention.eps}
% % (a) \\
% % \includegraphics[width=0.94\textwidth]{channel_attention.eps} \\
% % \fbox{\includegraphics[width=0.95\textwidth]{channel_attention.eps}}\\
% % (b)
% % \end{tabular}
% \caption{Illustration of the channel attention convolution layer.}
% \label{fig:attention}
% \end{wrapfigure}

% In channel dimension, we adopt another 1D convolution layer with kernel size $K_c$ that operates on the average map after applying a 2D average pooling function in spatial and temporal dimension, as demonstrated in the Fig \ref{fig:attention} (b), where $K_c$ is selected as $3$ in this work. Note that the 1D convolution layer muss have the same number of input channels and output channels, which can be different in the spatial attention layer.

In temporal dimension, we follow the multi-scale temporal modeling module from the work~\cite{chen2021channel}, i.e.,  operating the temporal information in four parallel branches of different dilated convolutional kernels, which have the same kernel size of $3 \times 1$.

\subsection{Attention based graph convolutional network}

Given defined spatial, temporal layers, an hybrid attention based graph convolutional block is formed. As shown in the Fig \ref{fig:block} (a), spatial ($Conv\_S$) and temporal ($Conv\_T$) layers are followed by a batch normalization layer (\emph{BN}) and a \emph{ReLU} activation function. A residual connection is added besides Spatial-Temporal block. 
In this work, we employ 10 basic blocks that have output channels with size $64, 64, 64, 64, 128, 128, 128, 256, 256, 256$, respectively. As demonstrated in Fig \ref{fig:block} (b), a \emph{BN} function is used at the beginning to normalize the input data. A global \emph{Average Pooling} (Avg\_pool) layer is adopted to pool feature map and reshape the feature maps to a uniform size. An additional dropout layer is adopted with drop rate $0.5$ to mitigate overfitting. At end of the network, a \emph{Softmax} is applied to do final prediction.

\begin{figure*}[h!]
    \begin{center}
    \addtolength{\tabcolsep}{-5pt}
    \renewcommand{\arraystretch}{0.1}
    \begin{tabular}{c|ccc|cc}
    \parbox[c]{0.12\textwidth}{Action name} & \multicolumn{3}{c|}{RGB images} & \parbox[c]{0.12\textwidth}{\ \ Hybrid mask}& \parbox[c]{0.12\textwidth}{\ \ \ Final mask} \\
    
    \parbox[c]{0.12\textwidth}{Watching time} &
    \parbox[c]{0.2\textwidth}{\includegraphics[width=0.2\textwidth]{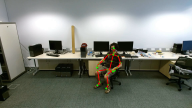}}&
    \parbox[c]{0.2\textwidth}{\includegraphics[width=0.2\textwidth]{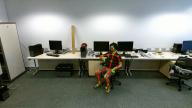}}&
    \parbox[c]{0.2\textwidth}{\includegraphics[width=0.2\textwidth]{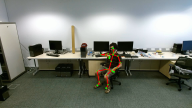}}&
    \parbox[c]{0.12\textwidth}{\includegraphics[width=0.12\textwidth]{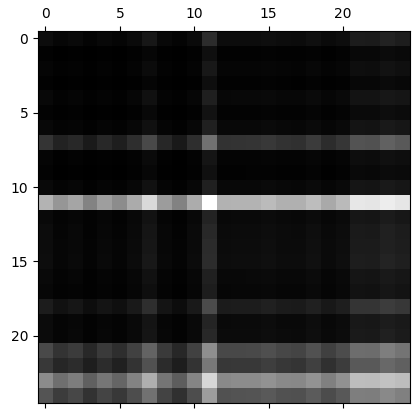}}&
    \parbox[c]{0.12\textwidth}{\includegraphics[width=0.12\textwidth]{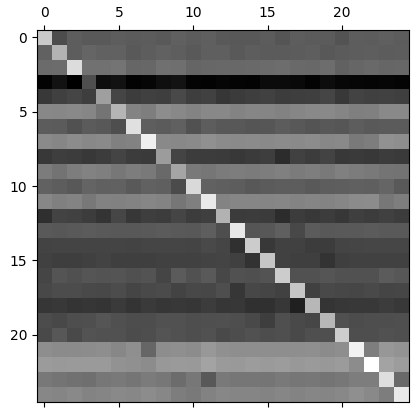}}\\
    \parbox[c]{0.12\textwidth}{Eating} & 
    \parbox[c]{0.2\textwidth}{\includegraphics[width=0.2\textwidth]{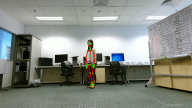}}&
    \parbox[c]{0.2\textwidth}{\includegraphics[width=0.2\textwidth]{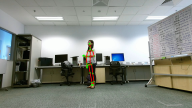}}&
    \parbox[c]{0.2\textwidth}{\includegraphics[width=0.2\textwidth]{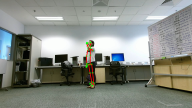}}&
    \parbox[c]{0.12\textwidth}{\includegraphics[width=0.12\textwidth]{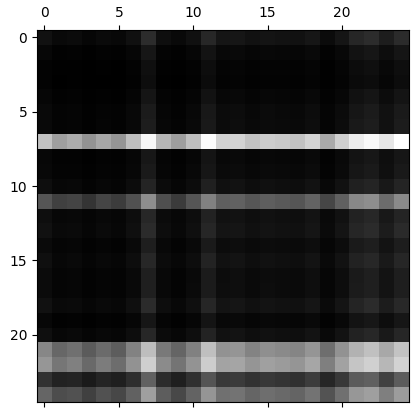}}&
    \parbox[c]{0.12\textwidth}{\includegraphics[width=0.12\textwidth]{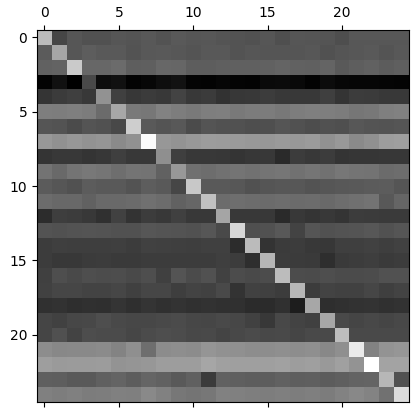}}\\
    \parbox[c]{0.12\textwidth}{Drinking} & 
    \parbox[c]{0.2\textwidth}{\includegraphics[width=0.2\textwidth]{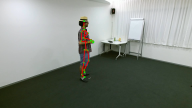}}&
    \parbox[c]{0.2\textwidth}{\includegraphics[width=0.2\textwidth]{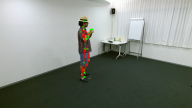}}&
    \parbox[c]{0.2\textwidth}{\includegraphics[width=0.2\textwidth]{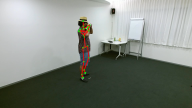}}&
    \parbox[c]{0.12\textwidth}{\includegraphics[width=0.12\textwidth]{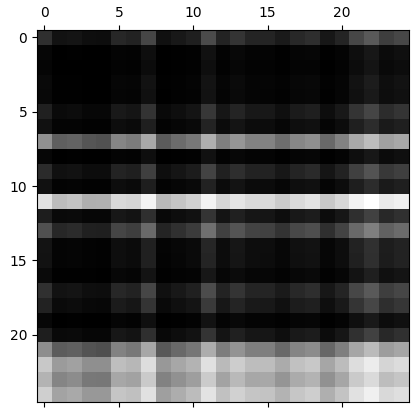}}&
    \parbox[c]{0.12\textwidth}{\includegraphics[width=0.12\textwidth]{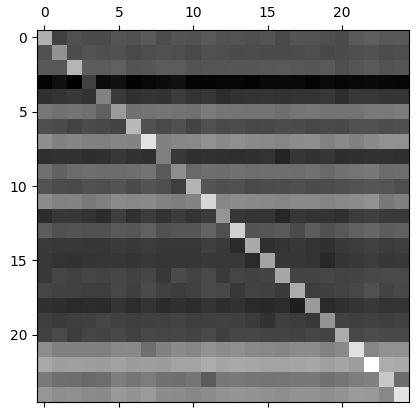}}\\
    
    \end{tabular}
    \caption{Qualitative results of hybrid attention mask and final identity output of 10-th spatial graph attention layer for different actions. Both have size of $25 \times 25$, where $25$ is number of skeleton joints.}
    \label{fig:mask}
    \end{center}
    \vspace*{-1.1\baselineskip}
\end{figure*}

\section{Experiments}
\label{sec:4}
To evaluate the performance of proposed HA-GCN, we experiment on large-scale human action recognition datasets: NTU-RGB+D~\cite{shahroudy2016ntu} and Kinetics human action dataset (Kinetics)~\cite{kay2017kinetics}.
We first perform detailed ablation study on the NTU-RGB+D cross-view benchmark to examine the contributions of the proposed model components to the recognition performance. Then, we evaluate the final model on both datasets and compare the results with other state-of-the-art methods.

\subsection{Dataset}
\textbf{NTU-RGB+D}~\cite{shahroudy2016ntu} is one of the largest and most challenging 3D action recognition dataset.
% which contains $56,000$ action clips with 3D skeleton data in $60$ action classes. The clips are performed by $40$ volunteers captured by 3 cameras  from different horizontal angles: $-45^\circ$, $0^\circ$ and $+45^\circ$. Each clip has at most $2$ subjects. 
In this work, we follow the recommend benchmarks: \textbf{cross-view} (X-View) and \textbf{cross-subject} (X-Sub).
% The authors of the dataset recommend two benchmarks: 1) \textbf{cross-view} (X-View) contains $37,920$ videos for training and $18,960$ videos for validation. The training set is captured by cameras $2$ and $3$, and the validation set is captured by camera $1$. 2) \textbf{cross-subject} (X-Sub) contains $40,320$ videos for training and $16,560$ videos for validation. Two sets are performed by different subjects. 

% In order to further examine the performance of the proposed model, we split the X-view dataset into body parts related (BPR) and pose related (PR) validation dataset, where the BPR action classes include: 1, 2, 3, 4, 6, 10, 13, 14, 15, 16, 17, 18, 19, 20, 21, 28, 34, 37, 38, 39, 40, 41, 44, 45, 46, 47, 48, and the rest is PR dataset.

\textbf{Kinetics}~\cite{kay2017kinetics} is a more challenging human action recognition dataset, which has $300,000$ videos in $400$ action classes retrieved from YouTube. 
In this work, we use 2D skeleton dataset ($240,000$ clips for training, $20,000$ clips for validating) that generated by Yan \textit{et al.}~\cite{yan2018spatial} using the OpenPose toolbox~\cite{openpose}.

% their released training skeleton dataset ($240,000$ clips) and report the top-1 and top-5 accuracy on the validation set ($20,000$ clips).
% Each clip in Kinetics lasts around 10 seconds. The dataset only provides raw video clips without skeleton information. 
% Y estimate the 2D skeleton information using the public OpenPose toolbox~\cite{openpose}. 
% At most two people are selected in the multi person videos based on the average joint confidence. The dataset is divided into a training set ($240,000$ clips) and a validation set ($20,000$ clips). 

\subsection{Configuration details}
All experiments are conducted on the PyTorch deep learning framework with single NVIDIA-2080ti GPU. The optimization strategy is stochastic gradient descent (SGD) with Nesterov momentum (0.9). $16$ batch size is used for training, $128$ batch size is applied for testing. Cross-entropy is adopted as the loss function for the back propagation. The weight decay is set to 0.0001. The full HA-GCN model has 1.42M parameters, and the RD- or RA-GCN has 1.34M parameters.

For NTU-RGB+D dataset, the size of input data is set the same as~\cite{shi2019two}, in which the max number of person in the samples is $2$, the longest frame is $300$. The learning rate is selected as $0.1$ and multiplied by $0.1$ at $60$-th and $90$-th epoch. The number of training epochs is set as $120$. 

For Kinetics dataset, we generate $150$ frames with $2$ persons in each sample. Then the data is slightly rotated and translated in random. It has the same learning rate and epoch setup as NTU-RGB+D dataset. 

In order to evaluate the effectiveness of the proposed components on different input streams, we introduce an improvement ratio metric as follows:
\begin{equation}
    r_{J/B} = \frac{Acc_{J} - Acc^*_{J}}{Acc_{B} - Acc^*_{B}}
\end{equation}
where $Acc$ and $Acc^*$ are accuracy of proposed model and baseline, respectively, $J$ means using joint input stream, and $B$ represent bone input stream. Assuming the improvement is positive, the proposed model is more promising for the joint input stream when $r_{J/B}>1$, and better for bone stream otherwise. For the negative improvement ratio, i.e., performance drop ratio, the removed component is more beneficial to the joint stream when $r_{J/B}>1$ and better for the bone stream otherwise.

\subsection{Ablation study}

\begin{table}
% \begin{wraptable}{r}{0.43\textwidth}
\centering
    % \begin{table}
        % \begin{center}
        % \vspace*{-\baselineskip}
            \caption{\label{tab:ablation}The performance in terms of accuracy of HA-GCN with RA attention and RD attention layer.}
            \begin{tabular}{|l|c|c|c|}
                \hline
                \multirow{2}{*}{Methods} & \multicolumn{2}{c|}{Accuracy} & Improvement ratio\\
                \cline{2-3} 
                & Joint stream& Bone stream & $r_{J/B}$\\
                \hline\hline
                AGCN$^*$ & $93.7\%$  & $93.2\%$ &$-$  \\
                AGCN  & $93.9\%$ & $93.5\%$ & $0.67$\\
                AGCN (plus)  & $95.0\%$ & $94.7\%$ & $0.86$\\
                RD-GCN & $95.6\%$ & $95.2\%$ & $0.95$\\
                RA-GCN  & $95.1\%$& $95.4\%$ & $0.64$\\
                % \rowcolor{cyan}
                HA-GCN (single T)& ${95.2}\%$ & ${94.9}\%$ & $0.94$\\
                HA-GCN (full)& $\mathbf{95.8}\%$ & $\mathbf{95.5}\%$ & $0.88$\\
                \hline
                HA-GCN (w/o RA) & $86.6\%$ & $93.6\%$ & $4.84$ \\
                HA-GCN (w/o RD) & $94.6\%$ & $85.2\%$ & $0.12$ \\
                \hline
            \end{tabular}
        
        % \end{center}
    % \end{table}
    \begin{tablenotes}
    \footnotesize
        \item[]$^*$ means using the original graph, and the rest experiments are conducted with new designed graph. AGCN is a single stream of 2s-AGCN. The AGCN ``plus" implements an additional convolutional layer after generating the attention graph. The ``single T'' HA-GCN uses the temporal convolutional layer from AGCN instead of 4 parallel dilated convolutional layers. The model without RA (w/o RA) turns off the RA branch in the test phase, and its improvement ratio (performance drop) is calculated in the respect to the full model, as the same as the model without RD branch(w/o RD).
    \end{tablenotes}
    \vspace*{-1.5\baselineskip}
% \end{wraptable}
\end{table}

We examine the contribution of proposed components to the recognition performance with the X-View benchmark on the NTU-RGB+D dataset. 

\textbf{Qualitative results:} The baseline is the single stream of 2s-AGCN (AGCN), which has $93.7\%$ and $93.2\%$ accuracy for joint and bone input stream, respectively. By using our new designed graph that has connection between head, hands and feet, they are slightly improved by $0.2\%$ and $0.3\%$. 
% The new graph has greater impact on the bone stream than joint stream, since the newly added edges actually increase the number of bones in the skeleton map. 
The following experiments are based on the new graph. 
Another interesting finding is that feeding the final graph mask into an additional convolutional kernel can significantly improve the prediction results. The improved results are list in the $3.$ row of Table~\ref{tab:ablation}, with an increase of $1.3\%$ and $1.5\%$ respectively over the original AGCN.

As introduced in Sec. \ref{method:attention_layer}, we have two types of attention mechanisms: RD and RA. We add them separately into the attention based graph convolution network. The performance of the RD-GCN and RA-GCN on different input streams are presented in the $4.$ and $5.$ rows of Table \ref{tab:ablation}. The result demonstrates that both of the RD and RA attention layers are beneficial to the skeletal action recognition performance. With both attention layers, it achieves the best performance, as shown in the last row of Table \ref{tab:ablation}. However, the independent contributions of RA and RD attentions to the different input streams are still unclear. The result can be also observed from the close improvement ratio $r_{J/B}$, which is $0.95$ for RD-GCN and $0.67$ for RA-GCN. This observation motivates to turn off one attention branch of the trained full model in the test phase and investigate how much performance drop this causes. As demonstrated in the last two rows of Table~\ref{tab:ablation}, we turn off RA (w/o RA) and RD (w/o RD), respectively, and examine top 1 accuracy on the X-View benchmark. It is clear that turning off RA attention branch results in more performance degradation in the joint stream than the bone stream ($9.8\%$ vs $1.9\%$) and turning off RD attention branch does the opposite, resulting in more performance drop in the bone stream ($2.2\%$ vs $10.3\%$). This result confirms that in the complete model, RD and RA attention mechanisms are more favorable for the bone stream and joint stream, respectively.
% in which the RD attention layer contributes slightly more to joint stream related action recognition whereas the RA attention seems promising for the bone stream related action classification.  More evidences can be found in quantitative results. 

\begin{table*}[h!]
    % \begin{center}
        \caption{\label{tab:nturgbd}Comparisons of accuracy ($\%$) with popular existing methods on the NTU-RGB+D cross-view and cross-subject dataset (Left, Top-1) and the Kinetics dataset$^1$ (Right, Top-1 and Top-5)}
        \begin{minipage}[t]{.48\linewidth}\centering
            \begin{tabular}{|l|cc|}
                \hline
                Methods & X-View & X-Sub \\
                \hline\hline
                Lie Group~\cite{vemulapalli2014human} & $52.8$ & $50.1$ \\
                % HBRNN~\cite{du2015hierarchical} &$64.0$  & $59.1$\\
                Deep-LSTM~\cite{shahroudy2016ntu} & $67.3$ & $60.7$\\
                VA-LSTM~\cite{zhang2017view} & $87.7$ & $79.2$\\
                TCN~\cite{kim2017interpretable} & $83.1$ & $74.3$\\
                Synthesized CNN~\cite{liu2017enhanced} & $87.2$ & $80.0$ \\
                $3$scale ResNet $152$~\cite{li2017skeleton} & $90.9$ & $84.6$ \\
                ST-GCN~\cite{yan2018spatial} & $88.3$ & $81.5$ \\
                $2$s AS-GCN~\cite{li2019actional} & $94.2$ & $86.8$ \\
                $2$s AGCN~\cite{shi2019two} & $95.1$ & $88.5$ \\
                $2$s AGC-LSTM~\cite{si2019attention} & $95.0$ & $89.2$ \\
                $4$s Directed-GNN~\cite{shi2019skeleton} & $96.1$ & $89.9$ \\
                $4$s Shift-GCN~\cite{cheng2020skeleton} & $96.5$ & ${90.7}$  \\
                $4s$ CRT-GCN~\cite{chen2021channel} & $96.8$ & $\mathbf{92.4}$ \\
                % $\drsh$ Single GPU & $95.2$ & $88.0$\\
                PoseC3D$^2$~\cite{duan2021revisiting} & $97.1$ & $94.1$ \\
                $\drsh$ Pose based ~\cite{duan2021revisiting} & $93.7$ & - \\
                \hline
                $1$s HA-GCN (ours) & $95.8$ & $89.4$\\
                $2$s HA-GCN (ours) & $96.6$ & $91.5$ \\
                $4$s HA-GCN (ours) & $\mathbf{ 97.0}$ & $92.1$ \\
                \hline
            \end{tabular}
            % \caption{Comparisons of the Top-1 accuracy (\%) with the state-of-the-art methods on the NTU-RGB+D cross-view dataset}
        \end{minipage}
        \begin{minipage}[t]{.48\linewidth}
        \centering                
            \begin{tabular}{|l|cc|}
                \hline
                Methods &  Top-1 (\%) & Top-5 (\%) \\
                \hline\hline
                Deep-LSTM~\cite{shahroudy2016ntu} &  $16.4$ & $35.3$\\
                TCN~\cite{kim2017interpretable} & $20.3$ & $40.0$\\
                ST-GCN~\cite{yan2018spatial} & $30.7$ & $52.8$ \\
                $2$s AGCN~\cite{shi2019two} & $36.1$ & $58.7$\\
                PoseC3D~\cite{duan2021revisiting} & $38.0$ & $59.3$\\
                \hline
                $1$s HA-GCN (ours) & $35.1$ & $58.0$\\
                $2$s HA-GCN (ours) & $37.4$ & $60.5$ \\
                $4$s HA-GCN (ours) & $\mathbf{38.2}$ & $\mathbf{61.1 }$ \\
                \hline
            \end{tabular}
            % \caption{Comparisons of the Top-1 and Top-5 accuracy (\%) with the state-of-the-art methods on the Kinetics dataset}
        \end{minipage}
        {\begin{tablenotes}
                \footnotesize
                    \item[a]$^1$ The pose data of Kinetics dataset is generated by OpenPose.
                    % \item[b]$^2$ The result is claimed by authors, the result in the next line is conducted under the same configuration on single GPU as other models.
                    \item[c]$^2$ The model used additional texture information, the pose based result is presented in the next line.

        \end{tablenotes}}
    % \end{center}

    \vspace*{-1.5\baselineskip}
\end{table*}

Besides the comparison in the spatial convolutional layer, we also conduct the comparison experiments with two types of the temporal convolutional layer, i.e., single convolutional layer and multi-scale temporal layer. The single convolutional layer is used in the ST-GCN \cite{yan2018spatial} and 2s-AGCN \cite{shi2019two}, which adopts a convolution kernel with a kernel size of $9\times1$ to extract features from adjacent frames in the temporal domain. The multi-scale temporal layer uses four parallel dilated convolutional kernels, which have different dilation size and wider receptive field in temporal dimension. The result of single temporal convolutional layer is listed in the $6.$ row of Table~\ref{tab:ablation} after HA-GCN (single T). Compared to the HA-GCN (single T), the full HA-GCN model using multi-scale temporal layer has significant performance improvement for both input streams.

\textbf{Qualitative results:} The attention mask and final output of $10$-th spatial hybrid attention layer for three action examples are visualized in the Fig~\ref{fig:mask}. In the example of \textit{watching time}, the subject is lifting his right forearm, which is also consistent with highlighting right hand joints ($11, 23, 24$) in the attention mask, whereas for the instance of \textit{eating} using left hand, the joints ($7, 21, 22$) belonging to the left hand have more weight in the attention mask. For the action using both hands, the position of two arms are highlighted in the attention mask, as demonstrated in the \textit{drinking} example (bottom) of the Fig~\ref{fig:mask}. The final identity output attention mask is demonstrated in right side of the Fig~\ref{fig:mask}, which is sum of attention mask and identity adjacent matrix.

\begin{figure}[t]
\vspace*{-2\baselineskip}
    \centering
    \vspace{2\baselineskip}
    \resizebox{\linewidth}{!}{
    \begin{tabular}{c c | c c}
    \multicolumn{4}{c}{
    \includegraphics[width=0.21\textwidth]{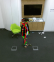}  \includegraphics[width=0.21\textwidth]{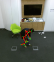}  \includegraphics[width=0.21\textwidth]{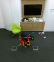}  \includegraphics[width=0.21\textwidth]{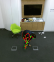} }\\
    \hline  
         \multicolumn{2}{c|}{Joint stream} & \multicolumn{2}{c}{Bone Stream} \\
         \includegraphics[width=0.2\textwidth]{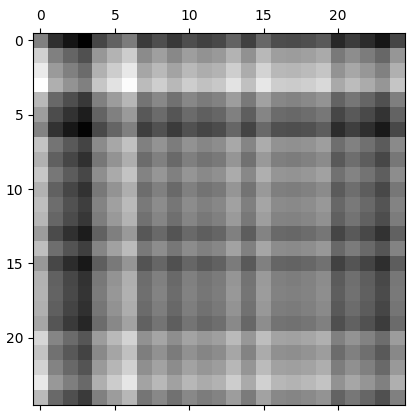} & \includegraphics[width=0.2\textwidth]{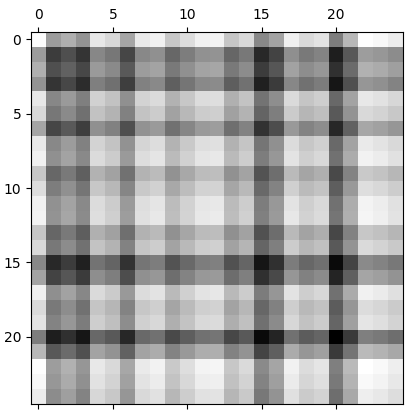} & \includegraphics[width=0.2\textwidth]{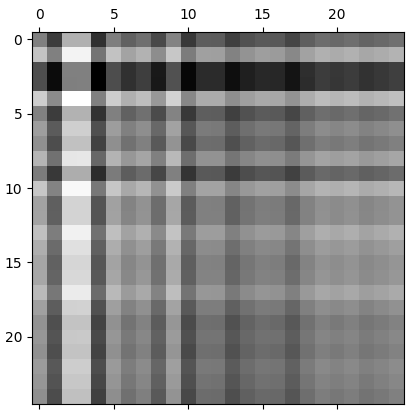} & \includegraphics[width=0.2\textwidth]{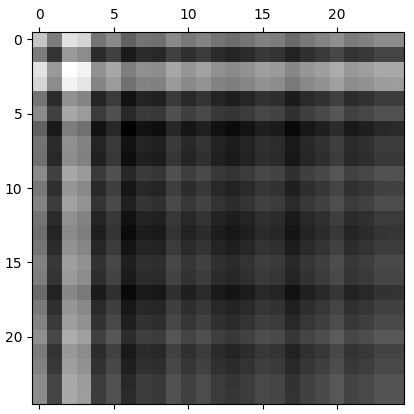} \\
         RD mask& RA mask& RD mask& RA mask\\
         
         Prediction: False & Prediction: True & Prediction): True & Prediction: False \\
         \includegraphics[width=0.2\textwidth]{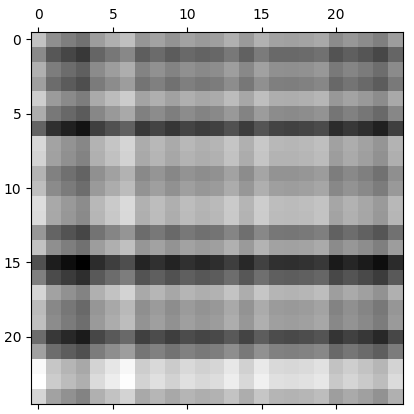} & \includegraphics[width=0.2\textwidth]{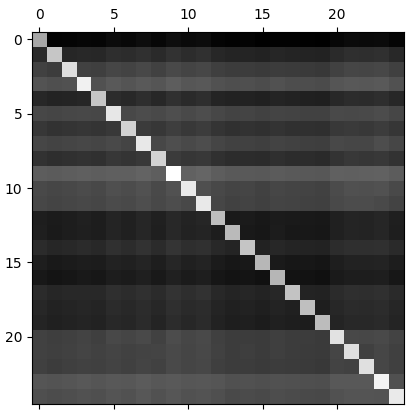} & \includegraphics[width=0.2\textwidth]{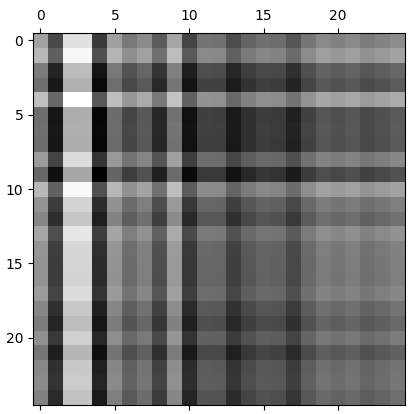} & \includegraphics[width=0.2\textwidth]{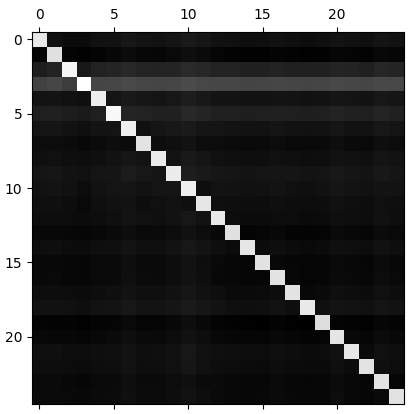} \\
         Hybrid mask& Final mask& Hybrid mask& Final mask\\
         \multicolumn{2}{c|}{Final Prediction: True} & \multicolumn{2}{c}{Final Prediction: True}\\
        %  \multirow{2}{*}{Joint stream}& RD & \includegraphics[width=0.2\textwidth]{figures/D_mask0_A1_9_0_cropped.png} & Ground-truth & Drinking \\
        %  & RA & \includegraphics[width=0.2\textwidth]{figures/D_mask0_A1_9_0_cropped.png} & Ground-truth & Drinking \\ \hline
        %  \multirow{2}{*}{Bone stream}& RD & \includegraphics[width=0.2\textwidth]{figures/D_mask0_A1_9_0_cropped.png}& Ground-truth & Drinking \\
        %  & RA & \includegraphics[width=0.2\textwidth]{figures/D_mask0_A1_9_0_cropped.png} & Ground-truth & Drinking \\
        %  & 
    \end{tabular}
    }
    \caption{Qualitative results of hybrid attention masks in the example of "putting on a shoe" on the joint and bone input streams. The top row displays the video images aligned with skeleton. The middle row presents the RD and RA attention masks for joint and bone input streams and their action prediction. The last row shows the hybrid and final attention mask and the final prediction.}
    \label{fig:qualitative}
    \vspace*{-1.5\baselineskip}
\end{figure}

In Fig.~\ref{fig:qualitative}, we present the hybrid attention masks and their predictions of the ``Putting on a shoe" example. It is obvious that RA and RD focus on different characteristics. 
In the joint stream in the Fig.\ref{fig:qualitative}, 
% RA shows strong correlation between nodes 0, 22, 23,and 24 (spine base, thumb left, hand tip right, thumb right), which leads a false prediction, 
RD shows strong correlation between nodes 1, 2, 3, 4, 5 and 6 (spine mid, neck, head, shoulder left, elbow left, wrist left), which is similar to the action of ``Taking off a shoe" and leads a false prediction, 
whereas RA is more concerned with whole body nodes and correct the prediction by mixing attention mask. 
In the bone stream, RA focus only the influence of nodes 2 and 3, while RD focus on the whole bone nodes and emphasizes nodes 2 and 3, and finally correct the prediction of the action class. These qualitative results further support the complementarity of the two attention mechanisms.

\subsection{Comparison with state of the art}

In order to verify the performance of the attention based model, we compare the final model with existing popular skeleton-based action recognition methods on both the NTU-RGB+D dataset and Kinetics dataset. The results of these two comparisons are respectively presented in Table \ref{tab:nturgbd}. The compared methods include 
Lie Group~\cite{vemulapalli2014human}, 
% HBRNN~\cite{du2015hierarchical}, 
Deep-LSTM~\cite{shahroudy2016ntu}, VA-LSTM~\cite{zhang2017view}, TCN~\cite{kim2017interpretable}, 
Synthesized CNN~\cite{liu2017enhanced}, 
$3$scale ResNet $152$~\cite{li2017skeleton}, ST-GCN~\cite{yan2018spatial}, $2$s AS-GCN~\cite{li2019actional}, $2$s AGCN~\cite{shi2019two}, $2$s AGC-LSTM~\cite{si2019attention}, $4$s Directed-GNN~\cite{shi2019skeleton}, $4$s Shift-GCN~\cite{cheng2020skeleton}, CRT-GCN~\cite{chen2021channel} and PoseC3D~\cite{duan2021revisiting}. $1s$ is only using joint data as the input. $2s$ means two streams that include joint and bone data. $4s$ is using four streams of input data, which are joint, bone, joint motion and bone motion, respectively. On both datasets, our model has a strong performance and the $4$ stream model outperforms previous state-of-the-art pure skeleton-based approaches on NTU RGBD X-View benchmark and Kinetics skeleton dataset. Note that the PoseC3D model uses not only the skeletal information, but also texture information, which is unfair to compare with other pure skeleton based methods. Hence, we list its performance of pose-based recognition on the Tab \ref{tab:nturgbd} as well.

The main reason of worse performance of all models on Kinetics Skeleton dataset is that the skeleton in the dataset contains only 2D skeletal information, while it is 3D in the NTU-RGBD dataset. It can be seen from this point that multi-dimensional information is very important for action recognition.

% \begin{table}
%     \begin{center}
%         \begin{tabular}{|l|cc|}
%             \hline
%             Methods &  Top-1 (\%) & Top-5 (\%) \\
%             \hline\hline
%             Deep-LSTM~\cite{shahroudy2016ntu} &  $16.4$ & $35.3$\\
%             TCN~\cite{kim2017interpretable} & $20.3$ & $40.0$\\
%             ST-GCN~\cite{yan2018spatial} & $30.7$ & $52.8$ \\
%             $2$s AGCN~\cite{shi2019two} & $36.1$ & $58.7$\\
%             \hline
%             $1$s AB-GCN (ours) & $34.3$ & $57.1$\\
%             $2$s AB-GCN (ours) & $36.6$ & $59.7$ \\
%             $4$s AB-GCN (ours) & $\mathbf{37.3}$ & $\mathbf{60.4 }$ \\
%             \hline
%         \end{tabular}
%     \end{center}
%     \caption{Comparisons of the Top-1 and Top-5 accuracy (\%) with the state-of-the-art methods on the Kinetics dataset}
%     \label{tab:kinetics}
% \end{table}

\section{Conclusion}
\label{sec:5}
In this work, we develop a novel hybrid attention based graph neural network (HA-GCN) with new designed graph for skeleton-based human action recognition. The hybrid mechanism can extract and combine corresponding attention information for different input streams. To this end, we propose a hybrid attention layer consisting of two branches: \textit{Relative Distance} and \textit{Relative Angle} attention. Two type of attention information are coupled by a trainable parameter in the spatial layer. The conducted experiments on two large-scale datasets demonstrate that our hybrid attention model can improve the performance of multi-streams skeletal action recognition. Furthermore, we slightly improve the initial adjacent matrix by connecting head, hands and feet. Since the interaction with environment is a part of action recognition task, and the interaction relations can be represented graphically, we are planning to extend the work to the field of human-object interaction recognition.

% We evaluate our model on two challenging human action recognition datasets: NTU-RGB+D and Kinetics skeleton dataset and the final model achieve the best performance on both of them. We find that the spatial attention layer can extract more interaction information for body parts related action prediction and the channel attention layer is more benefit for pose related action classification. The result that the proposed HA-GCN works well on both datasets proves the effectiveness of the proposed attention mechanism on graph convolution networks and the resulting HA-GCN model. We plan to extend this work to the field of human-object interaction recognition, which can also be represented graphically.

%  \section*{Acknowledgement} This work is supported by the funding of the Lighthouse Initiative Geriatronics by StMWi Bayern (Project X, grant no. 5140951) and LongLeif GaPa GmbH (Project Y, grant no. 5140953).
\subsection*{Acknowledgement}
This work is supported by the funding of the Lighthouse Initiative Geriatronics by StMWi Bayern (Project X, grant no. 5140951) and LongLeif GaPa GmbH (Project Y, grant no. 5140953).

\bibliographystyle{IEEEtran}
\bibliography{main}


% \end{thebibliography}

\end{document}